\documentclass[runningheads]{llncs}
\usepackage[T1]{fontenc}
\usepackage{graphicx}
\usepackage{amsfonts}
\usepackage{amsmath}
\usepackage{hyperref}
\usepackage{hyperref}
\hypersetup{
    colorlinks=true,
    linkcolor=blue,
    citecolor=blue,
    urlcolor=blue
}
\usepackage{color}

\begin{document}
\title{A Joint 2D–3D Statistical Shape Model for Orthopedic Reconstruction}
%
%
\author{Florence Dell'Aniello Picard\inst{1,2}\and
Pranav Poudel\inst{1,2} \and Nairouz~Shehata\inst{1,2} \and
Frédéric~Lavoie\inst{3} \and Herve~Lombaert\inst{1,2}} 

\authorrunning{F. Dell'Aniello Picard et al.}
%
\institute{Polytechnique Montréal, Canada\\
\email{florence.dellaniello-picard@polymtl.ca}
\and
Mila - Quebec AI Institute, Canada
\and
CHUM - University of Montreal Hospital, Canada\\
}
\maketitle              
\begin{abstract}
Three-dimensional femoral reconstruction from radiographs supports surgical planning, implant sizing, and post-operative follow-up, but remains ill-posed as X-ray projections discard depth information. Existing methods often incorporate a 3D statistical shape model (SSM) as a shape prior to guide reconstructions toward anatomically plausible shapes, relying on iterative 3D-to-2D projection matching. Yet, these approaches are computationally expensive and constrain their SSM to a single dimensionality, leaving the statistical relationship between 2D observations and 3D geometry largely unexploited and unexplored. We instead propose a joint 2D–3D SSM that explicitly captures the co-variation between 2D and 3D segmentations in a shared latent space. During training, 2D and 3D segmentations are registered to a common 3D template and its corresponding 2D projections, and the resulting stationary velocity fields are jointly decomposed using principal component analysis (PCA). This joint modeling allows the 2D-to-3D mapping to be learned directly from data rather than computing correspondences at inference time. For unseen subjects, the 3D shape is recovered directly by lifting the 2D latent coordinates to the 3D PCA subspace, thereby eliminating the need for iterative 3D-to-2D projection. Experiments on NMDID demonstrate that the proposed joint 2D–3D SSM outperforms a widely-used 3D-only SSM baseline while achieving inference approximately 4 times faster, at under 3 seconds per subject. The code is available at: \url{https://github.com/florence-dellaniello-picard/joint2d3d-ssm}.

\end{abstract}

\section{Introduction}
Fast and accurate patient-specific three-dimensional (3D) femoral reconstruction is highly sought in knee arthroplasty for pre-operative planning, implant sizing, and post-operative follow-up~\cite{HA2024100822,karade3DFemurModel2015,nolte3DShapeReconstruction2023}. Radiographs are widely accessible and routinely acquired in clinical practice, making them a practical imaging modality for reconstruction. However, reconstructing 3D geometry from two-dimensional (2D) X-ray images is fundamentally ill-posed: depth information is irreversibly lost during the 3D-to-2D projection, such that multiple distinct shapes can produce nearly identical radiographs~\cite{dengPixel2Voxel3DReconstruction2026}. Although computed tomography (CT) and magnetic resonance imaging (MRI) resolve this ambiguity by providing volumetric data, they entail higher clinical burden, including radiation exposure for CT and longer acquisition times for MRI~\cite{dengPixel2Voxel3DReconstruction2026,hussainModernDiagnosticImaging2022}. The development of methods that rapidly and accurately recover 3D geometry from radiographs alone is therefore of clinical interest.

To regularize the underdetermined 2D-to-3D reconstruction, existing methods mostly rely on statistical shape models (SSMs), which encode \textit{a priori} knowledge of population-level anatomical variation~\cite{cootesActiveShapeModelstheir1995,Gu_3DDX_MICCAI2024,reynekeReview2D3D2019}. By capturing a mean shape and its principal modes of variation, SSMs guide reconstructions to anatomically plausible configurations. SSMs are based on a single geometric dimensionality, for example, 2D-only or 3D-only, and built either from anatomical landmarks~\cite{adamsCanPointCloud2023,BHALODIA2024103034}, or deformation fields~\cite{bonarettiImagebasedVsMeshbased2014,rueckertAutomaticConstruction3D2003}. The former defines explicit correspondences across subjects, while the latter describes how each patient differs from a common template~\cite{reynekeReview2D3D2019}. Although the landmark-based approach is usually preferred for its simplicity and interpretability, deformation-based SSMs establish denser correspondences and offer a richer, more continuous representation of shape variation across a population~\cite{rueckertAutomaticConstruction3D2003,xuImage2SSMReimaginingStatistical2023}. A popular choice is to parameterize the deformations as stationary velocity fields (SVFs), which can be exponentiated to produce smooth, invertible deformations and are well-suited for principal component analysis (PCA)-based modeling due to their linearity in the Lie algebra~\cite{bonarettiImagebasedVsMeshbased2014}. For these reasons, this work adopts deformation-based SSMs, though the proposed method could be extended to landmark-based representations.

Conventionally, the 2D-to-3D reconstruction problem is formulated as an optimization problem: SSM parameters are iteratively adjusted to minimize a discrepancy between 3D  projections onto 2D planes and observed radiographs. Features such as anatomical landmarks~\cite{asvadiBoneSurfaceReconstruction2021}, contours~\cite{BAKA2011840}, or pixel intensities~\cite{luThreedimensionalSubjectspecificKnee2021} are used, with at least two views required to constrain the 3D shape~\cite{BAKA2011840,reynekeReview2D3D2019,zheng2D3DCorrespondence2009}. Each iteration requires projecting 3D features into 2D, which is computationally inefficient. Repeated across multiple iterations and views, this renders 2D-to-3D reconstruction time-consuming~\cite{karade3DFemurModel2015}. However, beyond this practical cost lies an even more fundamental limitation: existing SSMs are constructed in a single geometric dimensionality, e.g., either 2D or 3D, and lack an intrinsic representation of the statistical relationship between 2D observations and 3D geometry. Each reconstruction relies on an explicit 3D-to-2D projection at every iteration, and the mapping from 2D to 3D must be recovered anew for each subject rather than learned once from data~\cite{BAKA2011840,karade3DFemurModel2015,rueckertAutomaticConstruction3D2003,zheng2D3DCorrespondence2009}.

To address both the computational cost of iterative optimization and the single dimensionality of existing SSMs, we propose a joint 2D-3D SSM that learns the co-variation between 2D and 3D SVFs directly from data. For each subject, biplanar projections are simulated from CT-derived segmentations, producing paired 2D and 3D shapes. Anatomical variation across three views~---~anteroposterior (AP), mediolateral (ML), and 3D~---~is represented as SVFs~\cite{vercauterenNonparametricDiffeomorphicImage2007}, and PCA is applied to their concatenation~\cite{lombaertJointStatisticsCardiac2013}. Since the 2D-to-3D relationship is encoded in a joint latent space, 3D reconstruction of unseen shapes reduces to inferring 2D latent coordinates from 2D observations, followed by a simple projection onto the 3D subspace, without per-subject optimization or explicit 3D-to-2D projection. This is well-suited for orthopedics, where radiographs are routinely acquired, and fast 3D reconstruction benefits surgical planning. We evaluate the proposed framework on femur reconstruction using the New Mexico Decedent Image Database (NMDID)~\cite{Edgar2020NMDID}. Our contributions are as follows:
\begin{itemize}
\item We propose a joint 2D-3D SSM over paired SVFs that explicitly encodes the co-variation between 2D and 3D segmentations in a shared latent space.
\item We formulate 2D-to-3D reconstruction as a closed-form inference problem: estimating latent coordinates from 2D observations and projecting them onto the 3D subspace, eliminating per-subject iterative optimization.
\item We validate on femoral data from NMDID~\cite{Edgar2020NMDID}, demonstrating that our joint 2D-3D SSM outperforms a widely-used traditional 3D SSM baseline.
\end{itemize}
\begin{figure}[!t]
    \centering
    \includegraphics[width=1\linewidth]{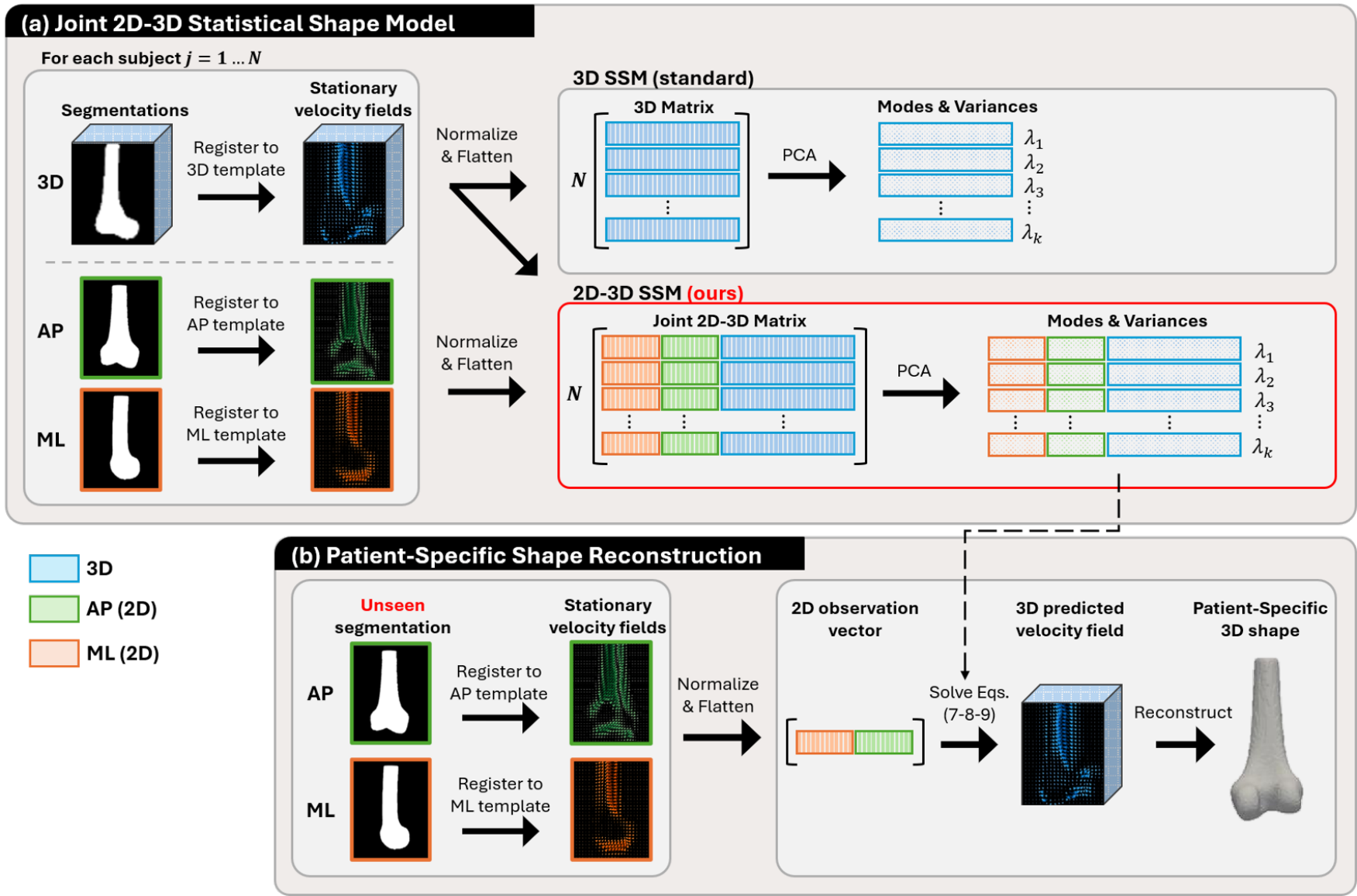}
    \caption{Overview of the proposed joint 2D–3D statistical shape model for patient-specific 2D-to-3D reconstruction. We learn a joint 2D-3D statistical shape model that captures co-variation between 2D and 3D images in a shared latent space, enabling fast and accurate 3D reconstruction from 2D images alone.}
    \label{fig:pipeline}
\end{figure}

\section{Methods}

Let us consider a dataset $\mathcal{D} = \left\{S_j^{2D}, S_j^{3D}\right\}_{j=1}^{N}$ of $N$ paired 2D and 3D segmentations, all rigidly registered to an arbitrarily selected reference frame. For each subject $j$, $S_j^{2D}=\left\{S_j^{(1)}, \ldots, S_j^{(P)}\right\}$ is a set of $P$ 2D segmentations from different imaging views, while $S_j^{3D}$ denotes the corresponding 3D segmentation. In our experiments, we use a standard clinical setting with $P=2$, and the views are the standard AP and ML projections. Our goal is to learn a joint 2D-3D SSM that couples 2D and 3D anatomical variation in a shared latent space, enabling direct estimation of $\hat{S}^{3D}$ from unseen $S^{2D}$ at inference time. Figure~\ref{fig:pipeline} illustrates the proposed framework, detailed below.

\subsection{Joint 2D-3D Statistical Shape Model} \label{methods:JSSM}
To capture patient-specific anatomical variations, we first define a common 3D reference template $\mathcal{T}^{3D}$, which can be any representative shape. In our experiments, we use the voxel-wise average of the 3D shapes in $\mathcal{D}$. We then diffeomorphically register each $S_j^{3D}$ to $\mathcal{T}^{3D}$, and each 2D segmentation $S_j^{(p)}$ to the corresponding 2D projection of the template, for $p=1,\ldots,P$. Each registration produces a deformation field, which we parameterize as an SVF, that maps the input 2D or 3D segmentations to their corresponding templates. As elements of a Lie algebra, SVFs exponentiate to diffeomorphic deformations, ensuring that any linear combination in latent space produces anatomically plausible shapes~\cite{arsignyLogeuclideanFrameworkStatistics2006}. For each subject $j$, we obtain a collection of SVFs:
${\mathbf{v}_j^{(1)}, \dots, \mathbf{v}_j^{(P)}, \mathbf{v}_j^{3D}}$, 
where $\mathbf{v}_j^{(p)}$ is the SVF for the $p$-th 2D view and $\mathbf{v}_j^{3D}$ is the SVF for the 3D view.

To construct the joint model, we flatten each SVF into a vector. For each view $m \in \mathcal{M}$, where $\mathcal{M} = \left\{1, \ldots, P, \text{3D}\right\}$ denotes the set of all 2D and 3D views, we independently z-score normalize the flattened SVF and scale it by the inverse square root of its dimensionality:

\begin{equation}
    \tilde{\mathbf{v}}_j^m = \frac{\mathbf{v}_j^m - \boldsymbol{\mu}_m}{\boldsymbol{\sigma}_m \sqrt{D_m}}, \quad m \in \mathcal{M}.
\end{equation}
In this expression, $\boldsymbol{\mu}_m$ and $\boldsymbol{\sigma}_m$ are the per-voxel mean and standard deviation of view $m$ computed over $\mathcal{D}$, and $D_m$ is the dimensionality of the flattened SVF of view $m$. The z-score normalization standardizes the distribution of each voxel across subjects, while dividing by $\sqrt{D_m}$ prevents higher-dimensional views from dominating the covariance, ensuring that each view contributes equally to the total variance of the joint representation.

We then stack all views of every subject into a shared data matrix $\mathbf{X} \in \mathbb{R}^{N \times D}$, where  $D=\sum_{p=1}^P D_{p} + D_{3D}$, such that the first dimensions of the paired SVF correspond to the 2D velocity fields, and the last dimensions to the 3D velocity fields. The $j$-th row of $\mathbf{X}$ corresponds to:
\begin{equation}
    \mathbf{x}_j = \left[\tilde{\mathbf{v}}_j^{(1)} \mid \cdots \mid \tilde{\mathbf{v}}_j^{(P)} \mid \tilde{\mathbf{v}}_j^{3D}\right].
\end{equation}

We apply PCA to $\mathbf{X}$, yielding principal components $\mathbf{u}_k$ and associated variances $\lambda_k$. Each component is partitioned as:
\begin{equation}
    \mathbf{u}_k = \left[\mathbf{u}_k^{(1)} \mid \cdots \mid \mathbf{u}_k^{(P)} \mid \mathbf{u}_k^{3D}\right].
\end{equation}
Any new multi-view shape can be approximated using the first $K$ principal components:
\begin{equation}
    \mathbf{x}_j \approx \sum_{k=1}^{K} z_{jk}\, \mathbf{u}_k,
\end{equation}
where $z_{jk}$ is the latent coordinate of subject $j$ along the $k$-th principal component, and $K$ is chosen to retain a desired percentage of the total variance.

\subsection{Patient-Specific Shape Reconstruction} \label{methods:inference}
Given the 2D segmentations $S_i^{2D}$ of an unseen subject $i$, we aim to recover the corresponding 3D shape $\hat{S}^{3D}$. We register each $S_i^{(p)}$ to their 2D projection of the template, obtaining 
SVFs $\mathbf{v}_i^{(1)}, \ldots, \mathbf{v}_i^{(P)}$, which are flattened, normalized and scaled following Eq.~(1) and concatenated into a 2D observation vector:
\begin{equation}
    \mathbf{x}_{i, obs} = \left[\tilde{\mathbf{v}}_i^{(1)} \mid \cdots \mid 
    \tilde{\mathbf{v}}_i^{(P)}\right].
\end{equation}

We seek the latent coordinates $\mathbf{z}$ that best explain the 2D observations while remaining close to the training distribution. Specifically, we minimize:
\begin{equation}
    \mathcal{L}(\mathbf{z}) = \left\|\mathbf{x}_{i,obs} - \mathbf{U}_{obs}^T \mathbf{z}\right\|^2 + \alpha\, \mathbf{z}^T \Lambda^{-1} \mathbf{z},
\end{equation}
where $\mathbf{U}_{obs} = \left[\mathbf{u}_k^{2D}\right]_{k=1}^{K}$ with $\mathbf{u}_k^{2D} = \left[\mathbf{u}_k^{(1)} 
\mid \cdots \mid \mathbf{u}_k^{(P)}\right]$ is the submatrix of principal 
components corresponding to the 2D views, $\Lambda = \mathrm{diag}(\lambda_1, 
\dots, \lambda_K)$ is the diagonal matrix of explained variances, and $\alpha > 0$ is a scalar controlling the strength of the regularization. Setting the gradient of $\mathcal{L}$ to zero yields the following closed-form linear system:

\begin{equation}
    \left(U_{obs} U_{obs}^T + \alpha \Lambda^{-1}\right)\mathbf{z} = U_{obs}\, \mathbf{x}_{i,obs},
\end{equation}
which is solved directly using Cholesky decomposition.

Once $\mathbf{z}$ is obtained, we reconstruct the 3D velocity field in the normalized and scaled space by projecting onto the 3D subspace of the principal components:
\begin{equation}
    \hat{\tilde{\mathbf{v}}}^{3D} = U_{3D}^T \mathbf{z},
\end{equation}
where $U_{3D} = \left[\mathbf{u}_k^{3D}\right]_{k=1}^{K}$. The result is then rescaled and denormalized to recover, in the original SVF space, the velocity field to be applied to the 3D template:
\begin{equation}
    \hat{\mathbf{v}}^{3D} = \left(\sqrt{D_{3D}}\,\hat{\tilde{\mathbf{v}}}^{3D}\right) \odot \boldsymbol{\sigma}_{3D} + \boldsymbol{\mu}_{3D},
\end{equation}
where $\odot$ denotes element-wise multiplication. 

Finally, $\hat{\mathbf{v}}^{3D}$ is exponentiated to a diffeomorphic deformation field using scaling and squaring~\cite{vercauterenNonparametricDiffeomorphicImage2007}. Since the deformation is estimated in the direction of the template, the inverse SVF is applied to $\mathcal{T}^{3D}$ to recover the predicted 3D shape:
\begin{equation}
    \hat{S}^{3D} = \mathcal{T}^{3D} \circ \exp\!\left(-\hat{\mathbf{v}}^{3D}\right),
\end{equation}
where $\exp(\cdot)$ is the exponential map, and $\circ$ spatial composition. Our entire approach thus requires only the 2D segmentations $S^{2D}$ as input to estimate $\hat{S}^{3D}$.

\section{Results}

Our experiments evaluate our proposed 2D-3D SSM for the reconstruction of 3D femoral shapes from DRRs of two standard biplanar clinical views, AP and ML. We first demonstrate that our method outperforms a standard 3D SSM baseline in reconstruction accuracy while achieving substantially reduced inference time. We further validate that the learned joint latent space captures meaningful and geometrically consistent modes of anatomical variation.

\subsection{Dataset and Preprocessing}
We use 1{,}368 CT scans of the lower limb from NMDID~\cite{Edgar2020NMDID}. The cohort comprises 781 individuals aged 7--97 years, of whom 230 are female and 551 are male. The dataset is randomly partitioned into four disjoint subsets: a \textit{registration set} (65\%), used to train VoxelMorph~\cite{balakrishnanVoxelMorphLearningFramework2019} for deformable registration; a \textit{shape-model set} (15\%), used to build the template and fit the joint 2D-3D SSM; a \textit{validation set} (10\%), used to select hyperparameters; and a held-out \textit{test set} (10\%), used solely for evaluation. When both femurs are available for the same individual, they are assigned to the same subset to prevent data leakage. To ensure consistency across the dataset, all CT scans are resampled to an isotropic spacing of $(1,1,1)$~mm. 

\subsubsection{2D and 3D Segmentations.} 3D femoral segmentations are automatically obtained using TotalSegmentator~\cite{wasserthalTotalSegmentatorRobustSegmentation2023}, removing the need for manual annotation. To establish a common coordinate frame for statistical shape modeling, right femurs are mirrored to the left side, and all segmentations are rigidly aligned to a reference femur using ANTs~\cite{tustisonANTsXEcosystemQuantitative2021}. The reference is selected as the first femur in the dataset. The aligned femurs are cropped to $180$~mm to ensure comparable length across subjects, and padded to a fixed volume size of $128 \times 256 \times 256$~vx. For each 3D segmentation, we simulate two standard clinical views ($P=2$), AP and ML, as DRRs of size $1024 \times 1024$~px using \textit{diffdrr}~\cite{gopalakrishnan2022fast}. DRRs are generated with a source-to-detector distance of 1800~mm, a pixel spacing of 0.304~mm, and a camera offset of $[0, 1350, 0]$~mm. We refer to this configuration as the \textit{projection setup}, and reuse it for all DRRs in this work. To obtain the 2D segmentations, DRRs are binarized.

\subsection{Experimental Setup} 
We implement our method in PyTorch~\cite{paszkePyTorchImperativeStyle2019} and run all experiments on a single NVIDIA RTX PRO 6000 Blackwell GPU.

\subsubsection{Registration Models.} Three VoxelMorph models~\cite{balakrishnanVoxelMorphLearningFramework2019} are trained on the \textit{registration set} for pairwise registration, one per view (AP, ML, and 3D), as each view operates in a different image space. Each model uses a U-Net encoder with $[32, 64, 128, 256]$ features and produces a diffeomorphic deformation via SVF integration (5 steps). Training minimizes a symmetric objective that combines a Dice loss between the two images being registered and a spatial gradient penalty on the velocity field, weighted by $0.2$. All models are optimized with Adam~\cite{kingma2014adam} using a learning rate of $10^{-4}$ for 5{,}000 iterations with a batch size of 8.

\subsubsection{Reference Templates.} The 3D template $\mathcal{T}^{3D}$ is constructed as the average of the \textit{shape-model set} segmentations. Its 2D counterparts are obtained by projecting $\mathcal{T}^{3D}$ using the AP and ML \textit{projection setup} and binarizing the result. This ensures that all three views are consistent projections of the same anatomy.

\subsubsection{Joint 2D--3D SSM.} Each femur in the \textit{shape-model set} is registered across all three views (3D, AP, and ML) to its respective template using the corresponding frozen VoxelMorph model. The resulting per-view SVFs are jointly decomposed using PCA, retaining components explaining 95\% of the variance ($K=38$). The regularization weight $\alpha = 10^{-5}$ is selected empirically on the \textit{validation set}.

\subsubsection{Baselines.} We compare our proposed method against a standard 3D SSM baseline that recovers the 3D femoral shape through iterative optimization~\cite{BAKA2011840,karade3DFemurModel2015,rueckertAutomaticConstruction3D2003,zheng2D3DCorrespondence2009}. This baseline is built by applying PCA to the same 3D SVFs used in our joint 2D-3D model, the only difference being that no 2D information is incorporated during training. We retain components explaining 95\% of the variance ($K=39$). At inference, the SSM shape parameters $\mathbf{z}$ are optimized with Adam~\cite{kingma2014adam} to align the soft-binarized DRR of the estimated shape with the signed distance field of the input AP and ML segmentations. A regularization term penalizes latent coordinates that deviate from the training distribution, encouraging anatomically plausible reconstructions. We use the \textit{projection setup} from the joint model, with an optimization step size of $10^{-2}$ and a regularization weight of $\alpha=0.2$. 

As a lower bound, we additionally report the mean shape, i.e.\ the 3D template $\mathcal{T}^{3D}$ returned regardless of input, reflecting the accuracy achievable without leveraging any 2D observations.

\subsubsection{Evaluation Metrics} 
Reconstruction quality is assessed on the held-out \textit{test set} using Dice similarity coefficient (DSC), 95th percentile Hausdorff distance (HD95), and mean surface distance (MSD) between the reconstructed and ground truth 3D femurs. We also report the average 2D-to-3D reconstruction time as a measure of computational efficiency.

\subsection{3D Reconstruction Evaluation} 

We assess whether the co-variation learned by the joint 2D-3D SSM translates into fast and accurate 3D reconstructions at inference. Table~\ref{tab:results} reports quantitative results on the held-out \textit{test set}. The joint 2D–3D SSM outperforms both the mean shape (lower bound) and the 3D SSM (standard) on every metric. It achieves a DSC of $96.15 \pm 1.65\%$, compared to $94.96 \pm 2.06\%$ for the 3D SSM and $91.25 \pm 5.50\%$ for the mean shape, the latter confirming that the 2D segmentations contribute meaningful subject-specific information. Surface-based metrics follow the same trend: MSD drops to $0.70 \pm 0.28$~mm versus $0.89 \pm 0.35$~mm for the 3D SSM baseline, and HD95 to $1.72 \pm 1.00$~mm versus $2.24 \pm 1.22$~mm. Beyond accuracy, our method completes inference in $2.76 \pm 0.09$~s, roughly four times faster than the 3D SSM ($11.08 \pm 0.90$~s), as the reconstruction is obtained in a single closed-form step rather than through iterative optimization. The remaining runtime is dominated by VoxelMorph registration and SVF integration.

\begin{table}[t]
\centering
\caption{Quantitative performance of the joint 2D-3D SSM (ours), the 3D SSM (standard), and the Mean Shape (lower bound) on 3D femur reconstruction. Results indicate average scores and standard deviations within the \textit{test set}. Bold indicates best results.}
\label{tab:results}
\resizebox{\linewidth}{!}{%
\begin{tabular}{l|c|c|c|c}
\hline
Method  & DSC (\%) $\uparrow$  & MSD (mm) $\downarrow$ & HD95 (mm) $\downarrow$   & Time (s) $\downarrow$ \\
\hline
Mean Shape (lower bound) &  91.25 $\pm$ 5.50 & 1.51 $\pm$ 1.03 & 3.81 $\pm$ 3.46  & 0.00 $\pm$ 0.00  \\
3D SSM (standard)   & 94.96 $\pm$ 2.06  & 0.89 $\pm$ 0.35 & 2.24 $\pm$ 1.22  & 11.08 $\pm$ 0.90 \\
Joint 2D-3D SSM (ours)  & \textbf{96.15 $\pm$ 1.65} & \textbf{0.70 $\pm$ 0.28}  & \textbf{1.72 $\pm$ 1.00}  & \textbf{2.76 $\pm$ 0.09}\\
\hline
\end{tabular}%
}
\end{table}

\begin{figure}[t]
    \centering
    \includegraphics[width=1\linewidth]{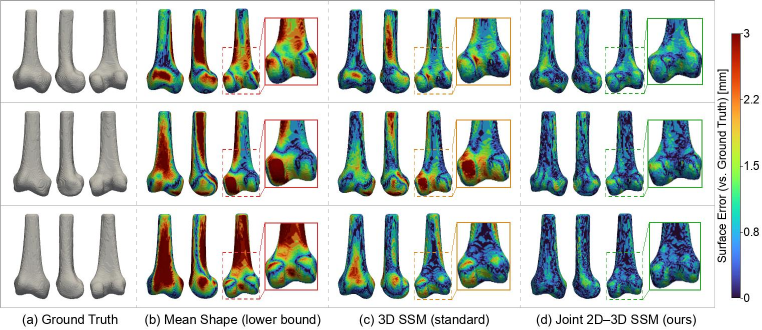}
    \caption{Surface reconstruction error for three representative test femurs, one per row. Colours indicate absolute surface distance to the ground truth (a) for the mean shape (lower bound) (b), the 3D SSM (standard) (c), and the proposed joint 2D-3D SSM~(ours)~(d). Values near zero indicate accurate reconstruction.}
    \label{fig:results_examples}
\end{figure}

To further analyze performance, Figure~\ref{fig:results_examples} presents the achieved reconstructions for three representative test femurs, each coloured by its surface distance to the ground truth. Our joint 2D-3D SSM consistently outperforms the 3D SSM baseline and the mean shape. The mean shape exhibits large errors throughout, reflecting its inability to adapt to individual anatomy. The 3D SSM substantially reduces these errors, yet they persist along the shaft and around the condyles. Our joint 2D-3D SSM produces reconstructions that are predominantly within 1~mm of the ground truth surface, with most pronounced gains at the condyles, which serve as key anatomical references for surgical planning~\cite{HA2024100822,karade3DFemurModel2015}.

\begin{figure}[t]
    \centering
    \includegraphics[width=1\linewidth]{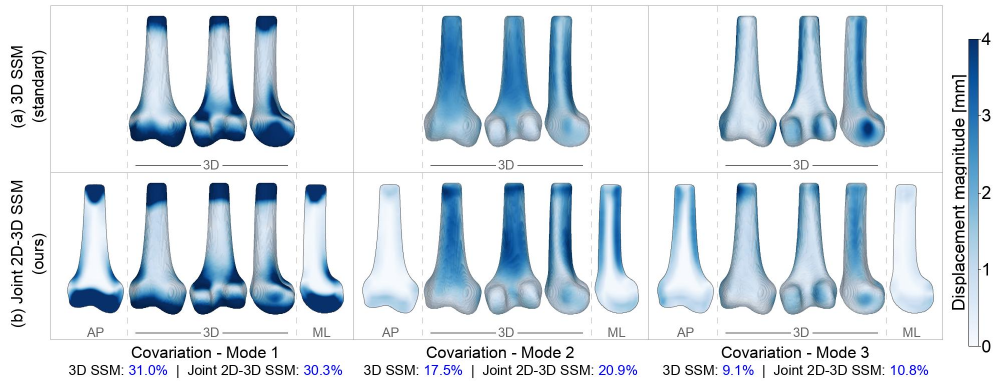}
    \caption{First three modes of variation of the 3D SSM (a) and the proposed joint 2D–3D SSM (b). Colours represent the local displacement magnitude (mm) at $\pm$2$\sigma$ from the mean shape, mapped on the mean geometry. Each mode is shown from three 3D orientations; for (b), AP and ML projection views are also included. The percentage of variance explained by each mode is reported at the bottom for both models.}
    \label{fig:modes}
\end{figure}

\subsection{Qualitative Analysis of the Latent Space} 
To validate that the learned latent spaces capture meaningful and coherent anatomical variation, we visualize and compare the modes of variation of the 3D SSM and the proposed joint 2D–3D SSM. For each mode, we colour the mean shape by local displacement magnitude at  $\pm$2$\sigma$ from the mean, providing an intuitive visualization of the regions most affected by each mode. Figure~\ref{fig:modes} presents three of these modes. 

Mode 1 captures variation predominantly at the distal end of the shaft and condyles, with large displacement magnitudes visible across all views for both the 3D SSM and the joint 2D-3D SSM. The consistent pattern across AP and ML views confirms that the joint 2D–3D SSM coherently captures this main source of variation. Mode 2 captures a more distributed variation along the shaft of the femur, with relatively high displacement magnitudes visible across all three 3D views. For the joint 2D–3D SSM, the ML view shows more pronounced deformation than the AP view, suggesting this mode primarily encodes mediolateral shape changes along the shaft. Mode 3 captures a more localized variation along the shaft and condyles, with deformation appearing more pronounced in the 3D SSM than in the joint 2D–3D SSM. This is reflected in the 2D projections, with the AP view presenting stronger displacements than the ML view.

Compared to the 3D SSM, the modes of the joint 2D–3D SSM are visually similar in 3D, suggesting that incorporating 2D views does not distort the learned 3D shape space. Furthermore, both models present a similar variance distribution: the first three modes account for 31.0, 17.5, and 9.1\% of the total variance for the 3D SSM, and 30.3, 20.9, and 10.8\% for the joint 2D–3D SSM. In total, 39 and 38 components are needed to explain 95\% of the total variance for the 3D SSM and the joint 2D–3D SSM, respectively, confirming that the joint model preserves the compactness of the 3D shape space while enriching it with a 2D-consistent structure. These results demonstrate that the proposed joint 2D–3D SSM learns a meaningful and compact latent space that coherently captures anatomical variation across both 2D and 3D representations.

\section{Conclusion}
We proposed a joint 2D–3D SSM that explicitly encodes the co-variation between 2D and 3D segmentations in a shared latent space. By coupling 2D and 3D segmentations at training time, 3D shape can be recovered from 2D segmentations alone through closed-form inference, requiring no iterative optimization. Experiments on NMDID demonstrate that our method outperforms a widely-used traditional 3D SSM on all metrics for 3D femur reconstruction: DSC improves from $94.96\%$ to $96.15\%$, HD95 from $2.24$~mm to $1.72$~mm, and MSD from $0.89$~mm to $0.70$~mm. Inference completes in $2.76$~s, roughly four times faster than the iterative 3D SSM baseline. Qualitative results show improvements at the condyles, a region of particular importance for surgical planning. Analysis of the learned latent space further confirms that the joint model captures meaningful and compact modes of anatomical variation, extending the 3D shape space with a 2D-consistent structure.

Our method currently assumes a known and fixed calibration. This assumption enables the simple and efficient linear reconstruction scheme we propose. Extending it to uncalibrated settings, for instance by jointly estimating imaging pose parameters alongside the shape latent variables, is a promising direction for future work that would broaden applicability to scenarios where calibration data is unavailable or patient positioning varies significantly. More broadly, the proposed 2D–3D SSM is not specific to femoral reconstruction. The underlying principle, learning a joint latent space from 2D projections and 3D volumes, can apply to any anatomical structure where paired 2D and 3D data are available, such as the hip, spine or cardiac imaging with fluoroscopy.

\subsubsection{Acknowledgment.}\label{sec:acknowledgment}
This work is supported by the Fonds de recherche du Québec (FRQNT), and by the Research Council of Canada (NSERC) through a Graduate Scholarship and an Alliance Advantage grant in partnership with Eiffel Medtech. Computational resources were partially provided by the Digital Research Alliance of Canada. The authors also thank the New Mexico Decedent Image Database (NMDID) for providing the CT scans used in this study.
 
%
%
%
\newpage
\bibliographystyle{splncs04}
\bibliography{references_LNCS}

@article{luThreedimensionalSubjectspecificKnee2021,
	title = {Three-dimensional subject-specific knee shape reconstruction with asynchronous fluoroscopy images using statistical shape modeling},
	journal = {Frontiers in Bioengineering and Biotechnology},
	author = {Lu, Hsuan-Yu and Shih, Kao-Shang and Lin, Cheng-Chung and Lu, Tung-Wu and Li, Song-Ying and Kuo, Hsin-Wen and Hsu, Horng-Chaung},
	year = {2021},
}

@article{asvadiBoneSurfaceReconstruction2021,
	title = {Bone surface reconstruction and clinical features estimation from sparse landmarks and {Statistical} {Shape} {Models}: a feasibility study on the femur},
	journal = {Medical Engineering \& Physics},
	author = {Asvadi, Alireza and Dardenne, Guillaume and Troccaz, Jocelyne and Burdin, Valérie},
	year = {2021},
}

@article{nolte3DShapeReconstruction2023,
	title = {{3D} shape reconstruction of the femur from planar {X}-ray images using statistical shape and appearance models},
	journal = {BioMedical Engineering OnLine},
	author = {Nolte, Daniel and Xie, Shuqiao and Bull, Anthony M. J.},
	year = {2023},
}

@article{tustisonANTsXEcosystemQuantitative2021,
	title = {The {ANTsX} ecosystem for quantitative biological and medical imaging},
	journal = {Scientific Reports},
	author = {Tustison, Nicholas J. and Cook, Philip A. and Holbrook, Andrew J. and Johnson, Hans J. and Muschelli, John and Devenyi, Gabriel A. and Duda, Jeffrey T. and Das, Sandhitsu R. and Cullen, Nicholas C. and Gillen, Daniel L. and Yassa, Michael A. and Stone, James R. and Gee, James C. and Avants, Brian B.},
	year = {2021},
}

@article{zheng2D3DCorrespondence2009,
	title = {A {2D}/{3D} correspondence building method for reconstruction of a patient-specific {3D} bone surface model using point distribution models and calibrated {X}-ray images},
	journal = {Medical Image Analysis (MedIA)},
	author = {Zheng, Guoyan and Gollmer, Sebastian and Schumann, Steffen and Dong, Xiao and Feilkas, Thomas and González Ballester, Miguel A.},
	year = {2009},
}

@inproceedings{xuImage2SSMReimaginingStatistical2023,
	title = {{Image2SSM}: {Reimagining} statistical shape models from images with radial basis functions},
	booktitle = {Medical {Image} {Computing} and {Computer} {Assisted} {Intervention} ({MICCAI})},
	author = {Xu, Hong and Elhabian, Shireen Y.},
	year = {2023},
}

@article{wasserthalTotalSegmentatorRobustSegmentation2023,
	title = {{TotalSegmentator}: {Robust} segmentation of 104 anatomic structures in {CT} images},
	journal = {Radiology: Artificial Intelligence},
	author = {Wasserthal, Jakob and Breit, Hanns-Christian and Meyer, Manfred T. and Pradella, Maurice and Hinck, Daniel and Sauter, Alexander W. and Heye, Tobias and Boll, Daniel T. and Cyriac, Joshy and Yang, Shan and Bach, Michael and Segeroth, Martin},
	year = {2023},
}

@inproceedings{vercauterenNonparametricDiffeomorphicImage2007,
	title = {Non-parametric diffeomorphic image registration with the demons algorithm},
	booktitle = {Medical {Image} {Computing} and {Computer} {Assisted} {Intervention} ({MICCAI})},
	author = {Vercauteren, Tom and Pennec, Xavier and Perchant, Aymeric and Ayache, Nicholas},
	year = {2007},
}

@incollection{paszkePyTorchImperativeStyle2019,
	title = {{PyTorch}: an imperative style, high-performance deep learning library},
	booktitle = {Advances in {Neural} {Information} {Processing} {Systems} ({NeurIPS})},
	author = {Paszke, Adam and Gross, Sam and Massa, Francisco and Lerer, Adam and Bradbury, James and Chanan, Gregory and Killeen, Trevor and Lin, Zeming and Gimelshein, Natalia and Antiga, Luca and Desmaison, Alban and Köpf, Andreas and Yang, Edward and DeVito, Zach and Raison, Martin and Tejani, Alykhan and Chilamkurthy, Sasank and Steiner, Benoit and Fang, Lu and Bai, Junjie and Chintala, Soumith},
	year = {2019},
}

@inproceedings{lombaertJointStatisticsCardiac2013,
	title = {Joint statistics on cardiac shape and fiber architecture},
	booktitle = {Medical {Image} {Computing} and {Computer} {Assisted} {Intervention} ({MICCAI})},
	author = {Lombaert, Hervé and Peyrat, Jean-Marc},
	year = {2013},
}

@article{karade3DFemurModel2015,
	title = {{3D} femur model reconstruction from biplane {X}-ray images: a novel method based on {Laplacian} surface deformation},
	journal = {International Journal of Computer Assisted Radiology and Surgery (IJCARS)},
	author = {Karade, Vikas and Ravi, Bhallamudi},
	year = {2015},
}

@article{HA2024100822,
	title = {{2D}-{3D} reconstruction of a femur by single x-ray image based on deep transfer learning network},
	journal = {IRBM},
	author = {Ha, Ho-Gun and Lee, Jinhan and Jung, Gu-Hee and Hong, Jaesung and Lee, HyunKi},
	year = {2024},
}

@article{cootesActiveShapeModelstheir1995,
	title = {Active shape models-their training and application},
	journal = {Computer Vision and Image Understanding (CVIU)},
	author = {Cootes, T.F. and Taylor, C.J. and Cooper, D.H. and Graham, J.},
	year = {1995},
}

@article{bonarettiImagebasedVsMeshbased2014,
	title = {Image-based vs. mesh-based statistical appearance models of the human femur: {Implications} for finite element simulations},
	journal = {Medical Engineering \& Physics},
	author = {Bonaretti, Serena and Seiler, Christof and Boichon, Christelle and Reyes, Mauricio and Büchler, Philippe},
	year = {2014},
}

@article{BHALODIA2024103034,
	title = {{DeepSSM}: {A} blueprint for image-to-shape deep learning models},
	journal = {Medical Image Analysis (MedIA)},
	author = {Bhalodia, Riddhish and Elhabian, Shireen and Adams, Jadie and Tao, Wenzheng and Kavan, Ladislav and Whitaker, Ross},
	year = {2024},
}

@article{BAKA2011840,
	title = {{2D}–{3D} shape reconstruction of the distal femur from stereo {X}-ray imaging using statistical shape models},
	journal = {Medical Image Analysis (MedIA)},
	author = {Baka, N. and Kaptein, B.L. and de Bruijne, M. and van Walsum, T. and Giphart, J.E. and Niessen, W.J. and Lelieveldt, B.P.F.},
	year = {2011},
}

@inproceedings{adamsCanPointCloud2023,
	title = {Can point cloud networks learn statistical shape models of anatomies?},
	booktitle = {Medical {Image} {Computing} and {Computer} {Assisted} {Intervention} ({MICCAI})},
	author = {Adams, Jadie and Elhabian, Shireen Y.},
	year = {2023},
}

@article{hussainModernDiagnosticImaging2022,
	title = {Modern diagnostic imaging technique applications and risk factors in the medical field: a review},
	journal = {BioMed Research International},
	author = {Hussain, Shah and Mubeen, Iqra and Ullah, Niamat and Shah, Syed Shahab Ud Din and Khan, Bakhtawar Abduljalil and Zahoor, Muhammad and Ullah, Riaz and Khan, Farhat Ali and Sultan, Mujeeb A.},
	year = {2022},
}

@inproceedings{arsignyLogeuclideanFrameworkStatistics2006,
	title = {A log-euclidean framework for statistics on diffeomorphisms},
	booktitle = {Medical {Image} {Computing} and {Computer} {Assisted} {Intervention} ({MICCAI})},
	author = {Arsigny, Vincent and Commowick, Olivier and Pennec, Xavier and Ayache, Nicholas},
	year = {2006},
}

@article{balakrishnanVoxelMorphLearningFramework2019,
	title = {{VoxelMorph}: a learning framework for deformable medical image registration},
	journal = {IEEE Transactions on Medical Imaging (T-MI)},
	author = {Balakrishnan, Guha and Zhao, Amy and Sabuncu, Mert R. and Guttag, John and Dalca, Adrian V.},
	year = {2019},
}

@article{rueckertAutomaticConstruction3D2003,
	title = {Automatic construction of 3-{D} statistical deformation models of the brain using nonrigid registration},
	journal = {IEEE Transactions on Medical Imaging (T-MI)},
	author = {Rueckert, D. and Frangi, A.F. and Schnabel, J.A.},
	year = {2003},
}

@article{reynekeReview2D3D2019,
	title = {Review of 2-{D}/3-{D} reconstruction using statistical shape and intensity models and x-ray image synthesis: {Toward} a unified framework},
	journal = {IEEE Reviews in Biomedical Engineering (RBME)},
	author = {Reyneke, Cornelius Johannes Frederik and Lüthi, Marcel and Burdin, Valérie and Douglas, Tania S. and Vetter, Thomas and Mutsvangwa, Tinashe E. M.},
	year = {2019},
}

@article{dengPixel2Voxel3DReconstruction2026,
	title = {{Pixel2Voxel}: {3D} reconstruction and visualization from limited number of x-rays with {3D}-aware diffusion models and iterative refinement},
	journal = {IEEE Transactions on Emerging Topics in Computing (TETC)},
	author = {Deng, Gaofeng and Ding, Sijie and Kaufman, Arie E.},
	year = {2026},
}

@inproceedings{kingma2014adam,
	title = {Adam: a method for stochastic optimization},
	booktitle = {International conference on learning representations ({ICLR})},
	author = {Kingma, Diederik P. and Ba, Jimmy},
	year = {2015},
}

@inproceedings{gopalakrishnan2022fast,
	title = {Fast auto-differentiable digitally reconstructed radiographs for solving inverse problems in intraoperative imaging},
	booktitle = {{MICCAI} {Workshop} on {Clinical} {Image}-based {Procedures} ({MICCAI}-{CLIP})},
	author = {Gopalakrishnan, Vivek and Golland, Polina},
	year = {2022},
}

@inproceedings{Gu_3DDX_MICCAI2024,
	title = {{3DDX}: {Bone} {Surface} {Reconstruction} from a {Single} {Standard}-{Geometry} {Radiograph} via {Dual}-{Face} {Depth} {Estimation}},
	booktitle = {Medical {Image} {Computing} and {Computer} {Assisted} {Intervention} ({MICCAI})},
	author = {Gu, Yi and Otake, Yoshito and Uemura, Keisuke and Takao, Masaki and Soufi, Mazen and Okada, Seiji and Sugano, Nobuhiko and Talbot, Hugues and Sato, Yoshinobu},
	year = {2024},
}

@misc{Edgar2020NMDID,
	title = {New mexico decedent image database},
	publisher = {Office of the Medical Investigator, University of New Mexico},
	author = {Edgar, Heather J. H. and Daneshvari Berry, Shamsi and Moes, E. and Adolphi, N. L. and Bridges, P. and Nolte, K. B.},
	year = {2020},
}
%





\end{document}